\documentclass[10pt,twocolumn,letterpaper]{article}

\usepackage[pagenumbers]{wacv} 
\definecolor{wacvblue}{rgb}{0.21,0.49,0.74}
\usepackage[breaklinks,colorlinks,allcolors=wacvblue]{hyperref}

\def\wacvPaperID{*****}
\def\confName{WACV}
\def\confYear{2026}

\title{Sparsity, Superposition, and Forgetting:\\
A Mechanistic Study of Representation Retention in Continual Learning}

\author{
  \makebox[0.45\textwidth][c]{
    \begin{tabular}{c}
      Jan Wasilewski\\
      Rochester Institute of Technology \\
Rochester NY, USA\\
      {\tt\small jw7630@rit.edu}
    \end{tabular}
  }
  \makebox[0.45\textwidth][c]{
    \begin{tabular}{c}
      Jedrzej Kozal\\
      Wrocław University of Science and Technology\\
      Wrocław, Poland \\
      {\tt\small jedrzej.kozal@pwr.edu.pl}
    \end{tabular}
  }\\[0.4cm] 
  \makebox[0.45\textwidth][c]{
    \begin{tabular}{c}
      Michał Woźniak\\
      Wrocław University of Science and Technology\\
      Wrocław, Poland \\
      {\tt\small michal.wozniak@pwr.edu.pl}
    \end{tabular}
  }
  \makebox[0.45\textwidth][c]{
    \begin{tabular}{c}
      Bartosz Krawczyk\\
      Rochester Institute of Technology \\
Rochester NY, USA\\
      {\tt\small bxkcis@rit.edu}
    \end{tabular}
  }
}

\begin{document}

\maketitle

\begin{abstract}
Continual learning (CL) systems often forget previously acquired knowledge, yet the mechanisms driving forgetting remain hard to isolate in practice because real datasets entangle many factors. We present a controlled, toy-world framework that makes these mechanisms observable and testable. Using a synthetic generator–separator pipeline, we define ground-truth latent “features,” build tasks with tunable sparsity and overlap, and introduce measurable quantities for representation strength and superposition (directional overlap among features). We then study retention dynamics—the temporal change of representation strength—by fitting sparse dynamical relations (via SINDy) between retention, superposition, and exposure history. A complementary task-level analysis based on effective rank characterizes how representational capacity is allocated across tasks. Our controlled experiments yield three takeaways. (1) Superposition tends to increase over time with transient dips at task boundaries, suggesting boundary-specific interference rather than steady drift. (2) Higher feature sparsity induces more superposition yet does not inevitably cause forgetting; when representations remain strong, forgetting can be reduced despite overlap. (3) Task-level effective rank grows with sparsity, indicating broader capacity usage under sparse regimes. Together, these results nuance the common intuition that more superposition leads to more forgetting by showing that overlap interacts with representation strength and capacity allocation. Our toy analysis provides falsifiable hypotheses and diagnostic tools for CL. 
\end{abstract}
\section{Introduction}
\label{sec:intro}

Artificial intelligence (AI) models deployed in non-stationary environments routinely face performance degradation. Continual learning (CL) aims to counter this by enabling models to acquire new skills without erasing prior knowledge, yet catastrophic forgetting remains a central obstacle: adapting to incoming tasks perturbs earlier representations and decision boundaries. While replay and regularization can mitigate forgetting, the underlying mechanisms—which features are overwritten, when, and why—are still not well understood. Prior evidence suggested that representational drift has only a limited impact on aggregate forgetting \cite{davari2022probingrepresentationforgettingsupervised}, reinforcing the view that changes in the classifier dominate \cite{wu2019largescaleincrementallearning}. More recent analyses challenge this picture, indicating that representational forgetting does occur but is harder to detect and quantify \cite{hess2024knowledge}. Complementary studies on the “tunnel effect” in CL further show that early layers disproportionately shape representation quality, out-of-distribution generalization, and forgetting, with top layers playing a smaller role \cite{masarczyk2023tunneleffectbuildingdata,harun2024variablesaffectoutofdistributiongeneralization}.

In parallel, mechanistic interpretability has highlighted how networks compress multiple concepts into limited resources via superposition—overlapping feature directions that arise when features co-occur sparsely. Superposition can be efficient, but may also increase interference across tasks. The resulting open question for CL is whether overlapping (polysemantic) features are intrinsically more prone to forgetting than disentangled ones, and how this depends on representation strength and capacity allocation. Addressing this requires isolating confounds that are ubiquitous in real data.

\noindent \textbf{Scope and approach.} We take a deliberately controlled route and introduce a toy-world framework that makes superposition and forgetting observable and testable. A synthetic generator–separator pipeline defines ground-truth latent “features,” tasks with tunable sparsity and overlap, and class labels via a fixed process. This setup lets us: (i) track single-feature dynamics—representation strength and overlap with other features; and (ii) study task-level allocation of latent capacity (effective rank) as tasks arrive. To connect dynamics with explanatory structure, we fit sparse differential relations (SINDy) that relate feature retention (temporal change of representation strength) to superposition and exposure history.

\noindent \textbf{Contributions.} Our work offers:
\begin{itemize}
\item A controlled experimental framework that jointly leverages ideas from mechanistic interpretability and continual learning, enabling precise measurement of superposition and retention without real-data confounds.
\item Formalizations of \emph{feature}, \emph{representation strength}, \emph{superposition}, \emph{feature retention}, and \emph{task embedding} into a unified dramework suited to sequential learning analysis.
\item Empirical evidence that sparser signals are represented with higher overlap and that superposition grows with new tasks; we also observe that sparser features tend to yield stronger representations.
\item A quantitative study of the interaction between representation strength, superposition, and retention; using SINDy, we find that features with weak representations under high overlap are especially vulnerable to forgetting.
\item A task-level analysis showing that tasks built from sparser features exhibit higher effective rank, indicating broader latent capacity usage.
\end{itemize}

\noindent \textbf{Findings at a glance.} In this controlled setting, we observe: (1) superposition generally rises over time with transient dips at task boundaries, suggesting boundary-specific interference rather than uniform drift; (2) higher feature sparsity induces more overlap, yet forgetting is not inevitable—features that maintain strong representations are more resistant even under high overlap; and (3) tasks composed of sparser signals occupy more latent “volume,” reflected by increased effective rank. These results nuance the common intuition that “more superposition means more forgetting” by showing that overlap interacts with representation strength and capacity allocation.

By cleanly separating causes from correlations in a toy world, our analysis surfaces falsifiable hypotheses and diagnostics for CL—tools that can guide principled algorithm design and targeted evaluations before scaling to complex, real-world settings.


\section{Motivation}

CL raises a mechanistic question that typical end-to-end evaluations cannot answer: \emph{which representations are overwritten, when, and why}? To study this, we need (i) a precise notion of a ``feature,'' (ii) a way to quantify when features are stored in overlapping directions (\emph{superposition}), and (iii) an experimental lens that isolates these factors from real-data confounds. This section motivates our choices along these three axes and lays the groundwork for the controlled setup in Sec.~3.

\subsection{What do we call a feature?}

Deep models extract intermediate signals that researchers variously call ``features.'' Dreyer et al. \cite{dreyer2024} treat the activations of deep neurons in CLIP models as features. Madry et al. \cite{NEURIPS2019_e2c420d9} adopt a broader definition, considering any function of the input as a feature. However, this definition presents challenges, as it does not naturally support semantically meaningful operations. Another perspective defines features as interpretable concepts \cite{elhage2022toymodelssuperposition}, but this approach is inherently constrained by human interpretability, potentially excluding useful representations that lack direct semantic meaning.

Because this work studies superposition, we adopt a definition of feature grounded in the linear representation hypothesis. Under this hypothesis, a feature corresponds to a direction, or more generally an approximately linear subspace, in a model's activation space. This geometric view is central to the toy models of superposition introduced by Elhage et al. \cite{elhage2022toymodelssuperposition}, where features are modeled as latent variables that the network represents using directions in a lower-dimensional hidden space. In that setting, superposition arises when the number of features exceeds the available representational dimensions and the model encodes multiple features using non-orthogonal directions. Thus, superposition is not simply a property of neurons or human-interpretable concepts; it is a property of the geometry of feature representations.

This distinction is important for our continual-learning analysis. We distinguish between \emph{ground-truth generative features} and \emph{learned representational features}. In our controlled toy world, a ground-truth feature is a coordinate of the sparse concept vector $f \in [0,1]^N$. The $i$-th feature can be isolated by the canonical basis vector $e_i$, and the generator maps this pure feature to an observable signal $\Phi(e_i)$. After training, we probe the encoder with this signal and define

$$
w_i(t) = E_\theta(\Phi(e_i)) \in \mathbb{R}^m,
$$

where $t$ indexes training time. Following the linear representation hypothesis, we interpret $w_i(t)$ as the learned representational direction associated with ground-truth feature $i$. This gives us a controlled bridge between known latent features in the data-generating process and their geometric realization inside the model.

\subsection{Why superposition and why measure it directly?}
Modern networks compress many potential features into limited latent capacity. When only a few features are active per input, overlapping directions can be an efficient coding strategy—superposition—where multiple concepts share non-orthogonal representations \cite{Olah2020ZoomIA}. While efficiency is beneficial, overlap also creates the potential for interference: updates that strengthen one direction may inadvertently perturb another. In realistic datasets, however, superposition is difficult to quantify because the feature set itself is unknown and entangled with data semantics. Our controlled setting removes this ambiguity by fixing the feature basis a priori. This lets us track (i) single-feature dynamics—how a feature’s representation strength evolves and how much it overlaps with others—and (ii) task-level allocation of latent capacity—how many effective directions are devoted to a collection of features. Together these views connect a mechanistic quantity (overlap) to a phenomenon of interest (forgetting) in a way that is otherwise inaccessible.

\subsection{Why the continual learning lens?}
In non-stationary streams, neural networks tend to lose performance on earlier tasks, a phenomenon known as catastrophic forgetting \cite{catastrophic_forgetting,Chen2018LifelongML}. CL studies how to learn sequentially while retaining past knowledge \cite{Chen2018LifelongML}. Among mitigation strategies, experience replay is both simple and widely used: it periodically revisits a small buffer of past samples (or their synthetic variants) during training \cite{DBLP:journals/corr/abs-1902-10486}. We focus on replay for two reasons. First, it provides a stable training protocol that minimizes confounds from stronger regularizers or architectural changes, allowing us to attribute observed effects to representation dynamics rather than to the mitigation method itself. Second, replay exposes a natural tension between efficiency and retention: as tasks accumulate, latent capacity pressure increases and superposition is expected to grow, making it an informative regime in which to study when overlap correlates with forgetting.

\section{Controlled Experimental Setup}

\begin{figure}[ht]
    \centering
    \includegraphics[width=0.9\linewidth]{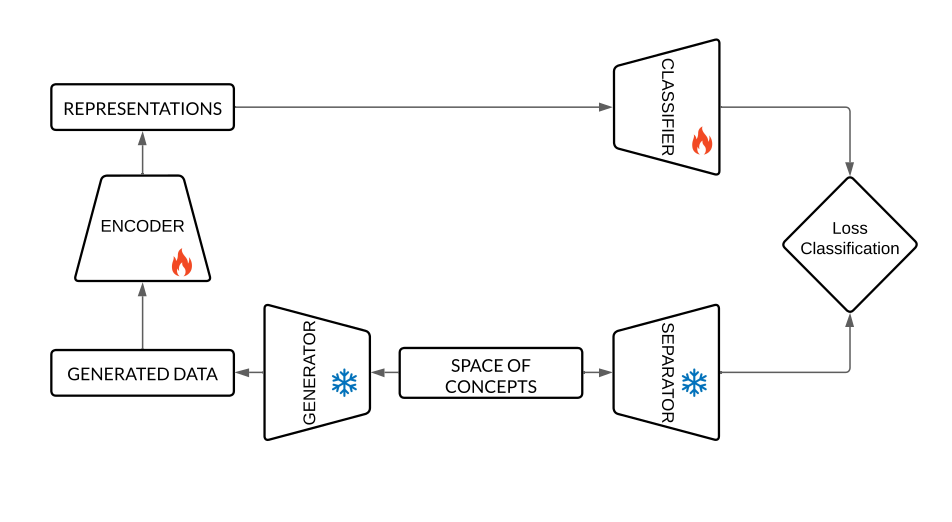}
    \caption{Diagram of the experimental setup. The \textbf{generator} and \textbf{separator} are randomly initialized frozen neural networks () that generate data sample and label. The \textbf{encoder}, \textbf{classifier}, and \textbf{decoder} modules are trainable (), and learning is performed sequentially with a replay buffer.}
    \label{fig:experiment_diagram}
\end{figure}

Measuring superposition in neural networks remains unsettled—there is no widely accepted, model-agnostic metric. We therefore introduce a controlled framework that lets us operationally define a \emph{feature}, its \emph{representation strength}, \emph{superposition strength}, a \emph{task embedding}, and \emph{feature retention}. The framework has two components: (i) \textbf{Data Generation}, a synthetic pipeline that produces labeled examples with known latent features and tunable sparsity/overlap; and (ii) \textbf{Neural Network Processing}, a replay-trained classifier whose internal representations we analyze. Figure~\ref{fig:experiment_diagram} summarizes the setup; formal definitions follow in Secs.~3.1--3.3.

\subsection{Data Generation}
\label{sec:data-gen}

\paragraph{Objects and notation.}
Let $N$ denote the feature (concept) dimension, $D$ the signal dimension, $K$ the number of classes, and $T$ the number of sequential tasks. Each datum is produced from a sparse nonnegative feature vector $\mathbf f\in[0,1]^N$ with support $\mathcal S(\mathbf f)=\{i:\,f_i>0\}$ and sparsity $\|\mathbf f\|_0\le s$.

\paragraph{Generator and separator.}
A fixed generator $\Phi:[0,1]^N\!\to\!\mathbb R^{D}$ maps features to an observable signal, and a fixed separator $\Psi:[0,1]^N\!\to\!\{1,\dots,K\}$ assigns labels \emph{in feature space}:
\begin{equation}
    X=\Phi(\mathbf f), \qquad y=\Psi(\mathbf f).
    \label{eq:gen-sep}
\end{equation}
In our instantiation, $\Phi$ and $\Psi$ are randomly initialized and kept \emph{frozen} throughout all experiments; $\Psi$ implements a predetermined linear rule followed by a softmax, while $\Phi$ is a feed-forward mapping. Features $\mathbf f$ are sampled by first drawing a support of size $s$ and then i.i.d.\ amplitudes on $[0,1]$ for indices in the support, with zeros elsewhere.

\paragraph{Tasks as distributions.}
Continual learning is simulated by presenting tasks $\{\mathcal D_t\}_{t=1}^T$ sequentially. Task $t$ activates a designated index set $\mathcal I_t\subseteq[N]$ and draws samples with $\mathcal S(\mathbf f)\subseteq \mathcal I_t$. Unless otherwise specified, the sets $\{\mathcal I_t\}$ are disjoint, yielding task-specific subspaces in feature space while leaving the generator and separator fixed across all tasks. Concrete architecture and sampling choices appear in the Appendix.

\subsection{Modeling}
\label{sec:modeling}

\paragraph{Network.}
The trainable model comprises an \emph{encoder} $E_\theta:\mathbb R^{D}\!\to\!\mathbb R^{m}$ and a \emph{classifier} $C_\theta:\mathbb R^{m}\!\to\!\{1,\dots,K\}$. We intentionally choose $m<|\mathcal I_t|$ (per-task concept count) to enforce a latent bottleneck and induce representational pressure. The overall predictor is $x\mapsto C_\theta\!\big(E_\theta(x)\big)$.

\paragraph{Training protocol.}
Tasks arrive in order $t=1,\dots,T$. We train with standard experience replay: a bounded buffer $\mathcal B$ stores a subset of past samples, and each update mixes the current task minibatch with items from $\mathcal B$. This provides a stable, widely used baseline and limits confounds introduced by stronger regularizers or architectural changes. Additional training details (optimizer, buffer size, schedules) are given in the Appendix.

\subsection{Single-feature perspective}
\label{sec:single-feature}

\begin{figure}[ht]
    \centering
\includegraphics[width=\linewidth]{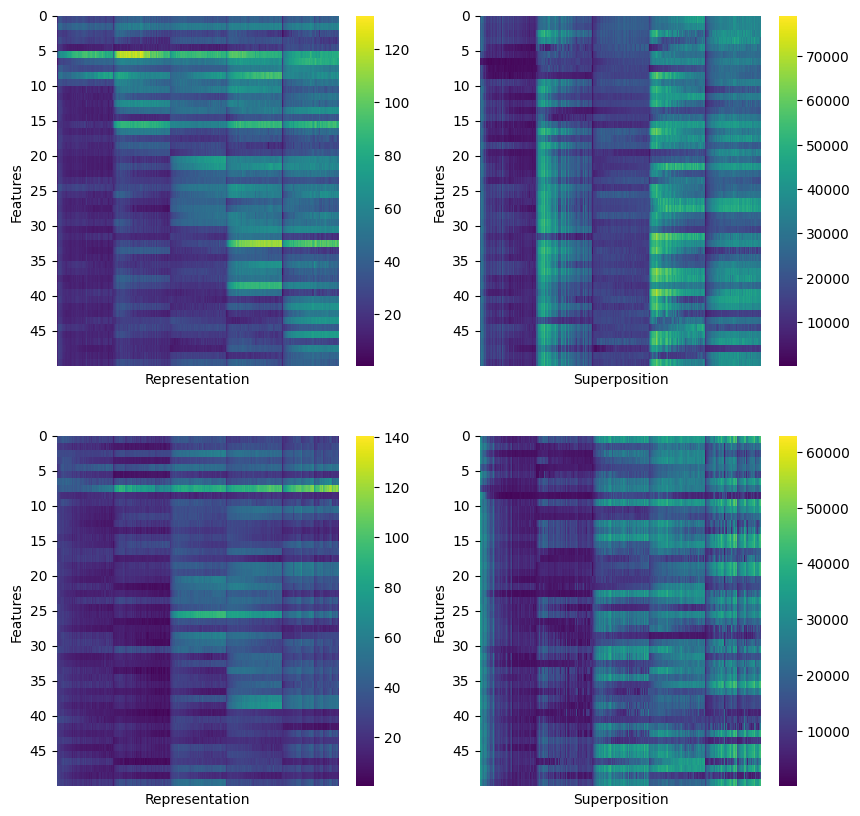}
    \caption{Heatmap of superposition and representation of the features evolution in time for feature sparsity 0.15 (top) and 0.2 (bottom)}
    \label{fig:heatmap}
\end{figure}
\begin{figure*}[ht]
    \centering
\includegraphics[width=\linewidth]{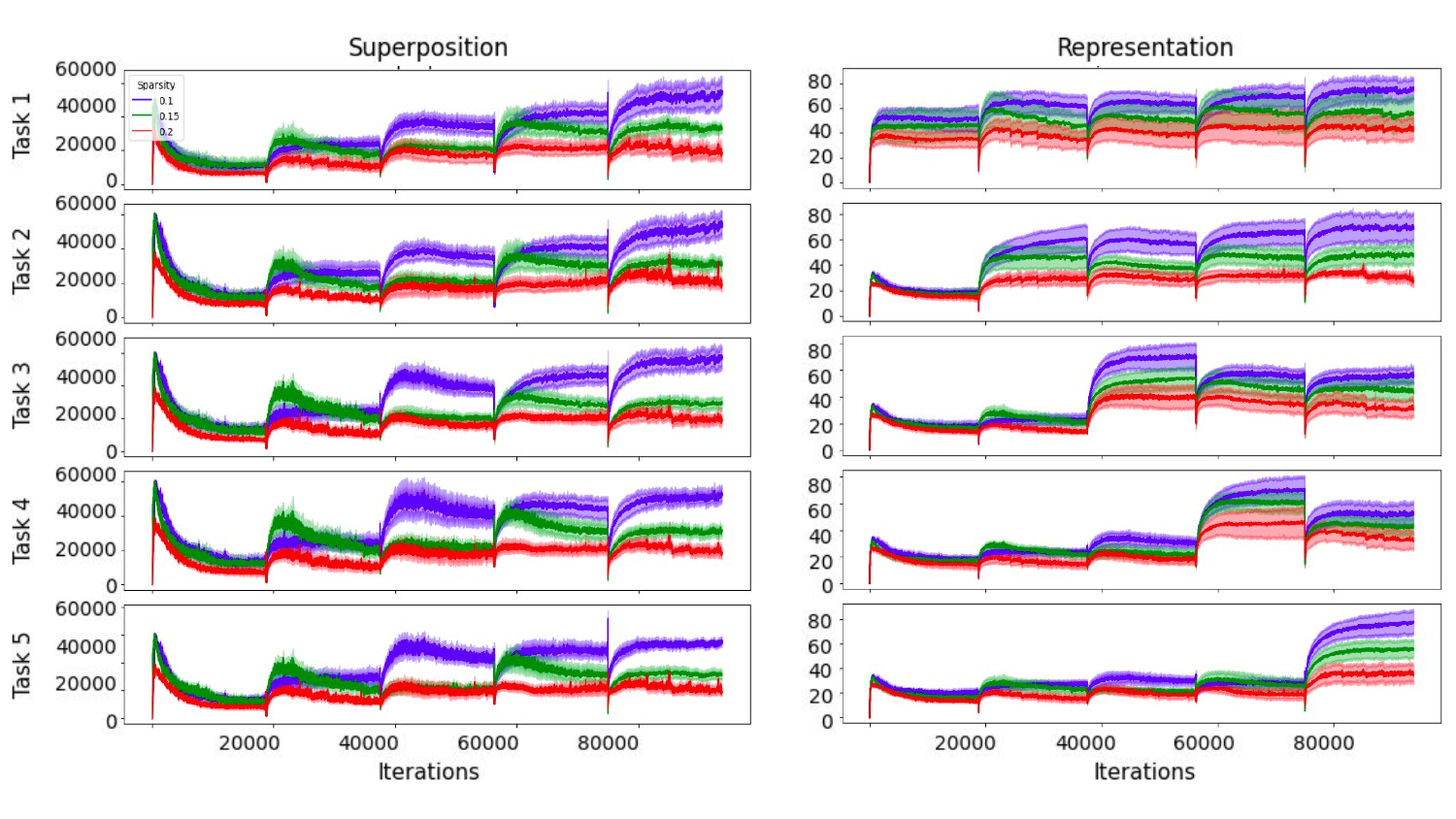}
    \caption{Dynamics of superposition and representation in time}
    \label{fig:cumulative_superposition}
\end{figure*}
\begin{figure*}[htb]
  \centering
  \includegraphics[width=\textwidth]{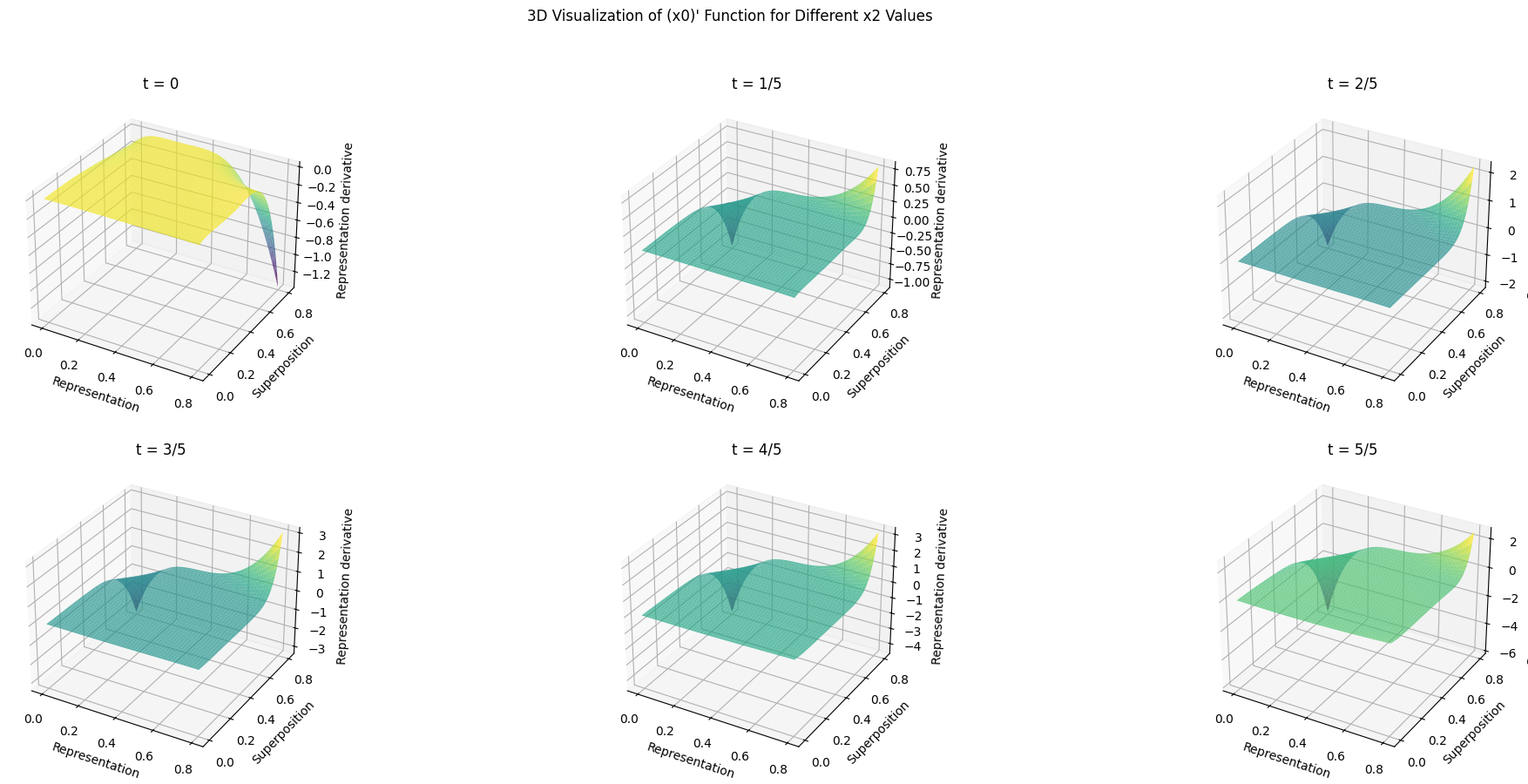}
  \caption{Evolution of feature retention as a function of representation strength and superposition.}
  \label{fig:sindy}
\end{figure*}

To make feature-level measurements well-posed, we probe the model with \emph{pure features}. This follows from the linear representation hypothesis introduced above: if a ground-truth feature is represented approximately linearly by the encoder, then activating that feature in isolation should reveal its corresponding direction in representation space. Let $\mathbf e_i$ denote the $i$-th canonical basis vector in feature space. The corresponding pure-feature signal is $\Phi(\mathbf e_i)$, and its \emph{feature embedding} at training time $t$ is
\[
    \mathbf w_i(t)\;=\;E_\theta\!\big(\Phi(\mathbf e_i)\big)\in\mathbb R^{m},
\]
where $t$ indexes training time (or updates). We define:

\paragraph{Representation strength.}
The \emph{representation strength} of feature $i$ at time $t$ is the embedding norm
\begin{equation}
    R(i,t)\;=\;\big\|\mathbf w_i(t)\big\|_2.
    \label{eq:repr-strength}
\end{equation}

\paragraph{Superposition strength.}
Let $\mathbf v_i(t)=\mathbf w_i(t)/\|\mathbf w_i(t)\|_2$ be the unit direction. The \emph{superposition strength} of feature $i$ aggregates how much other features project onto $\mathbf v_i(t)$:
\begin{equation}
    S(i,t)\;=\;\sum_{j\ne i}\big(\mathbf w_j(t)^\top \mathbf v_i(t)\big)^2.
    \label{eq:superposition}
\end{equation}
Intuitively, large $S(i,t)$ indicates that $i$ shares its latent direction with many other features (overlap), while small $S(i,t)$ suggests relative disentanglement.

\paragraph{Feature retention.}
We study \emph{retention dynamics} via the temporal derivative of representation strength:
\begin{equation}
    F(i,t)\;=\;\frac{\partial}{\partial t}\,R(i,t).
    \label{eq:retention}
\end{equation}
Our goal is to relate $F(i,t)$ to quantities available from the single-feature probe, namely $R(i,t)$, $S(i,t)$, and the time since feature $i$ first appeared in training, $t_{\mathrm{since}}(i,t)$. Concretely, we posit
\begin{equation}
    \frac{\partial R}{\partial t}\;=\;g\!\big(R,S,t_{\mathrm{since}}\big),
    \label{eq:sindy-target}
\end{equation}
and identify $g$ \emph{from data} using Sparse Identification of Nonlinear Dynamical Systems (SINDy)~\cite{Brunton_2016}.

\paragraph{Estimating $g$ with SINDy.}
We construct a library of candidate interactions (polynomials up to degree 10), standardize the inputs $(R,S,t_{\mathrm{since}})$, and estimate the sparse coefficients with thresholded least squares. To avoid leakage from untrained features, we apply an upper-triangular mask aligned to feature/task introduction times so that derivatives are fit only where the feature has been observed. The learned differential relation reveals how superposition and representation jointly govern retention/forgetting.

\subsection{Task-related set of features perspective}
\label{sec:task-view}

We next move from individual features to \emph{task embeddings}. For task $t$, we consider two complementary constructions:

\begin{enumerate}
\item \textbf{Pure-feature stack.} Stack embeddings of pure features for indices in $\mathcal I_t$: 
$T_t^{\text{pure}}(t')=[\,\mathbf w_i(t')\,]_{i\in\mathcal I_t}\in\mathbb R^{m\times |\mathcal I_t|}$.
\item \textbf{Data-driven stack.} Stack embeddings of observed samples from task $t$ at time $t'$:
$T_t^{\text{data}}(t')=[\,E_\theta(\Phi(\mathbf f^{(n)}))\,]_{n:\,(\mathbf f^{(n)},y^{(n)})\sim\mathcal D_t}$.
\end{enumerate}

These matrices summarize the region of latent space currently allocated to task $t$.

\paragraph{Task volume via effective rank.}
We quantify the \emph{volume} a task occupies with the \emph{effective rank}~\cite{eff_rank} of its embedding matrix $T$:
\begin{equation}
    \mathrm{EffRank}(T)\;=\;\exp\!\Big(-\sum_{i} p_i \log p_i\Big),
    \qquad
    p_i=\frac{\sigma_i}{\sum_j \sigma_j},
    \label{eq:effrank}
\end{equation}
where $(\sigma_i)$ are the singular values of $T$. Effective rank captures the number of energetically significant directions (spectral entropy) and is robust to small singular values: low values indicate heavy compression/overlap (consistent with strong superposition), whereas higher values indicate more directions devoted to representing the task. We track $\mathrm{EffRank}\!\big(T_t^{\text{pure}}(t')\big)$ and $\mathrm{EffRank}\!\big(T_t^{\text{data}}(t')\big)$ over time to study how latent capacity is reallocated as new tasks arrive.

\begin{figure}[ht]
    \centering
\includegraphics[width=1\linewidth]{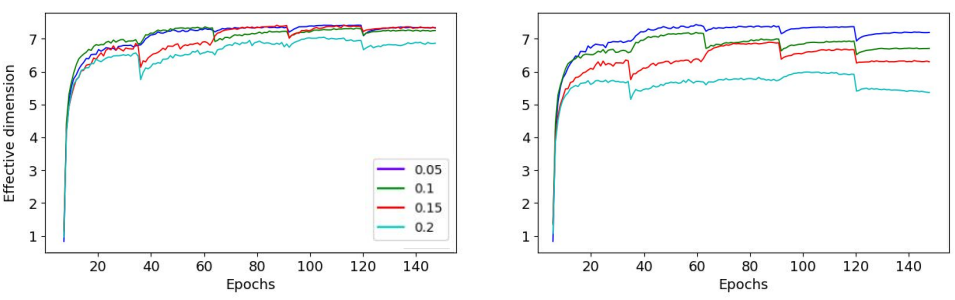}
    \caption{Sum of effective dimensions of all the tasks in time vs sparsity}
    \label{fig:sum_eff_dim}
\end{figure}

\begin{figure*}[htb]
  \centering
  \includegraphics[width=\textwidth]{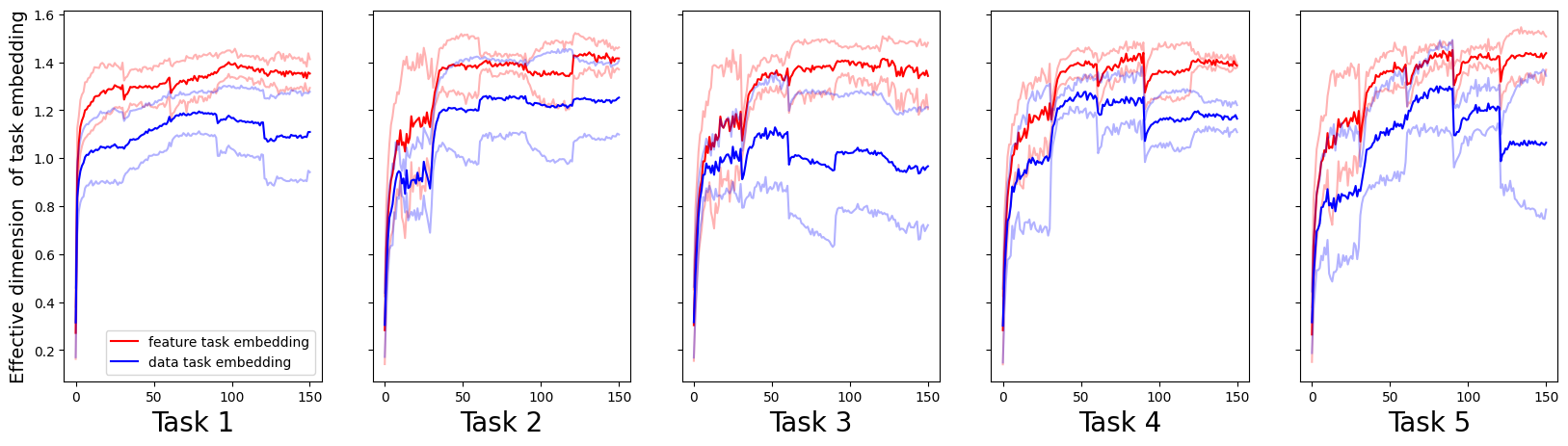}
  \caption{Evolution of effective dimensions of tasks embeddings. The task is mapped into region with similar effective rank no matter if the model was trained on it or not.}
  \label{fig:eff_dim_per_task}
\end{figure*}

\section{Experiments}
\label{sec:experiments}

In this section we address three research questions that guide our analysis:

\begin{itemize}
    \item \textbf{RQ 1} \emph{Do sparser feature signals produce more superposition in continual learning?} 
    \item \textbf{RQ 2} \emph{Are features stored in superposition more susceptible to forgetting than features with more disentangled representations?}
    \item \textbf{RQ 3} \emph{What is the relation between feature sparsity and the effective dimension of task embeddings in sequential learning (i.e., how the model allocates latent capacity across tasks)?}
\end{itemize}

Unless otherwise noted, the architectures and training protocol follow Table~1 and Sec.~3. Additional settings and standard performance metrics (e.g., task-wise accuracies over time) are included in the Appendix.

\medskip
\noindent\textbf{RQ 1: Sparsity vs. superposition over time.}
Figure~2 visualizes the joint evolution of \emph{superposition strength} $S(i,t)$ and \emph{representation strength} $R(i,t)$ for pure features across sparsity regimes, and Figure~3 shows their dynamics over time. Three consistent observations emerge:
\begin{enumerate}
    \item \emph{Sparser features exhibit higher overlap.} As sparsity increases, features are encoded with more directional sharing in the latent space, reflected by systematically larger $S(i,t)$ in Figure~2. This indicates that limited latent capacity is economized via overlap when only a few features are active per input.
    \item \emph{Superposition rises gradually but dips at task boundaries.} Across settings, Figure~3 shows a characteristic pattern: a \emph{pronounced transient dip} in $S(i,t)$ when a new task starts, followed by a \emph{gradual recovery and continued growth}. The boundary dip is consistent with boundary-specific interference and rapid re-alignment as the model accommodates new features, while the subsequent growth reflects slow consolidation of shared directions.
    \item \emph{Representation and superposition co-evolve.} Features with stronger representations $R(i,t)$ tend to stabilize into persistent directions even as $S(i,t)$ increases. The heatmaps reveal that high-$R$ regions are not necessarily low-$S$; instead, they tend to occupy \emph{intermediate-to-high} $S$ with reduced volatility, suggesting that overlap and strength are \emph{not} strictly antagonistic.
\end{enumerate}
Overall, these patterns indicate that sparsity pushes the network toward overlapping codes, but the time course of superposition is structured: boundary-induced reconfiguration followed by consolidation.

\medskip
\noindent\textbf{RQ 2: Are overlapping features more prone to forgetting?}
We study \emph{feature retention} $F(i,t)=\partial_t R(i,t)$ as a function of $R(i,t)$, $S(i,t)$, and the time since first exposure $t_{\mathrm{since}}(i,t)$. Figure~4 summarizes the learned dynamical relation using SINDy (Eq.~\eqref{eq:sindy-target}) and yields the following insights:

\begin{enumerate}
    \item \emph{Vulnerability concentrates in the low-$R$, high-$S$ regime.} The most negative retention (largest drops in $R$) consistently occurs when a feature is \emph{weakly represented} yet \emph{highly overlapped}. Intuitively, weak signals that share a direction are easily overwritten by updates driven by other features occupying the same subspace.
    \item \emph{Strong representations buffer the effect of overlap.} As $R$ increases, the dependence of $F$ on $S$ becomes less adverse and can even invert locally (flat or slightly positive slopes in the high-$R$ band). This indicates that overlap alone does not determine forgetting; \emph{representation strength} is a critical moderator.
    \item \emph{Time-since-first-seen acts as a consolidation variable.} Conditioning on $(R,S)$, $t_{\mathrm{since}}$ reduces volatility in $F$—early after first exposure, $F$ is more negative in high-$S$ regions (boundary fragility), whereas later it becomes closer to zero (or slightly positive) as shared directions stabilize. This aligns with the \emph{dip-then-recover} behavior at task onsets observed in Figure~3.
\end{enumerate}

Taken together, the results nuance the common intuition “more superposition $\Rightarrow$ more forgetting.” Overlap increases vulnerability \emph{when} representations are weak and fresh; once a feature has accumulated sufficient strength, its direction becomes more resilient even within a shared subspace.

\medskip
\noindent\textbf{RQ 3: How does sparsity affect task-level capacity allocation?}
We examine task embeddings using the \emph{effective rank} (Eq.~\eqref{eq:effrank}). Figure~5 tracks the \emph{sum} of effective ranks across tasks over time; Figure~6 reports per-task effective rank.

\begin{enumerate}
    \item \emph{Sparser regimes increase effective rank.} Both the aggregate (Figure~5) and per-task (Figure~6) curves rise with sparsity, indicating that sparser signals occupy a broader set of latent directions. This suggests the network allocates more “volume” to encode tasks built from sparser features, even as feature-level overlap increases.
    \item \emph{Task-wise rank is surprisingly uniform across seen and unseen tasks (early on).} Figure~6 shows that, at the moment a new task arrives, the effective rank for its embedding region is comparable to that of previously seen tasks. This \emph{early uniformity} implies that the encoder maintains a relatively stable latent scaffold into which new features are mapped, before fine-grained adjustment occurs as training proceeds.
    \item \emph{Boundary effects mirror feature-level dynamics.} Consistent with RQ~1, we observe small rank contractions at task transitions followed by recovery. This “contraction–expansion” behavior suggests transient compression of directions to accommodate new features, then re-spreading as representations consolidate.
\end{enumerate}

\medskip
\noindent\textbf{Additional settings and robustness.}
We evaluate alternative settings in the Appendix (Figures~7–10). The qualitative conclusions above persist across these variants: (i) sparsity increases superposition, with boundary dips and gradual recovery; (ii) low-$R$/high-$S$ regions concentrate retention loss, while high-$R$ features are comparatively robust; and (iii) effective rank increases with sparsity and exhibits consistent boundary transients. We also report SINDy stability analyses (Figures~11–12), noting that while coefficient selection can vary across runs, the \emph{sign structure} of the dominant interactions (negative dependence on $S$ at low $R$, moderating with $R$ and $t_{\mathrm{since}}$) is stable and aligns with the empirical dynamics seen in Figures~2–3.


\section{Key Findings}

\noindent\textbf{What we learned.} Our controlled study reveals a structured relationship between sparsity, overlapping representations, and forgetting. First, as inputs activate fewer ground–truth features, the encoder systematically economizes latent capacity via increased \emph{superposition}. This growth is not monotone: superposition exhibits transient dips at task boundaries followed by consolidation–driven recovery. Second, overlap is \emph{conditionally} harmful. The most vulnerable regime combines \emph{low representation strength} with \emph{high overlap}; once a feature accumulates sufficient strength, its embedding remains comparatively stable even within shared subspaces. Third, at the task level, sparser regimes occupy more latent “volume,” reflected by higher effective rank of task embeddings, indicating broader allocation of directions to encode task structure despite greater feature–level sharing. Taken together, these findings refine the common intuition that “more superposition $\Rightarrow$ more forgetting,” highlighting representation strength and capacity allocation—not overlap alone—as the primary moderators of retention in continual learning.

\noindent\textbf{Why a toy world—and why it matters.} Real data entangle concept frequency, co–occurrence statistics, and label structure, making it difficult to attribute forgetting to specific representational mechanisms. Our toy generator–separator pipeline provides ground–truth features, tunable sparsity/overlap, and a fixed labeling rule, which jointly (i) make superposition and retention \emph{measurable}, (ii) enable time–resolved probes of single–feature dynamics (strength, overlap, retention) and task–level capacity (effective rank), and (iii) surface \emph{falsifiable} hypotheses—e.g., boundary dips, low–$R$/high–$S$ vulnerability—that can be tested on richer architectures and datasets. This controlled lens complements conventional accuracy curves by exposing \emph{how} internal representations stretch, compress, and stabilize as tasks arrive, thereby turning vague narratives about “interference” into concrete, testable mechanisms.

\noindent\textbf{Impact for the CL community.} Mechanistic summaries like ours help translate representation dynamics into actionable guidance. The results suggest capacity–aware evaluation protocols that track effective rank alongside accuracy; retention–oriented diagnostics that flag low–$R$/high–$S$ regions as early–warning indicators of forgetting; and algorithmic hypotheses (e.g., prioritizing consolidation of weak, overlapped features; scheduling replay to smooth boundary dips; or explicitly budgeting latent directions per task). More broadly, the framework provides a reusable set of probes—pure–feature embeddings, overlap metrics, and effective–rank traces—that can be layered onto standard CL benchmarks to advance a mechanistic understanding of the interplay between internal representations and forgetting in neural networks.

\section{Related works}
\label{sec:related_works}
\subsection{Continual learning}

Continual Learning \cite{Chen2018LifelongML} is a Machine Learning area, where data distribution is not stationary. Without access to prior tasks catastrophic forgetting occurs \cite{catastrophic_forgetting}. Several studies have tried to explain how representations change over time when we train in a Continuous setting. In \cite{davari2022probingrepresentationforgettingsupervised} it was shown that the effect of forgetting in representations is minimal even without any forgetting prevention. This led to several works, that focused primarily on changing classifier \cite{wu2019largescaleincrementallearning,chrysakis2023online} to address forgetting. Recent study \cite{hess2024knowledge} challenges these findings by examining linear probe accuracy before and after training with a given task. When we compare this accuracy, we can find that forgetting in representations does take place. Rypeść et al. \cite{NEURIPS2024_73ba81c7} studies the shift in representations during knowledge distillation from the past model. Their findings seem to align with ours, as they show that the rank of the covariance matrix of classes gradually rises with each task. 

This work focused on the setting based on experience replay \cite{DBLP:journals/corr/abs-1902-10486}, as it is the most popular method for catastrophic forgetting prevention \cite{Chen2018LifelongML}. The impact of feature representation on training continually with rehearsal was studied before. In \cite{caccia2022newinsightsreducingabrupt} the impact of abrupt representation change after introducing new classes is examined. 

\subsection{Mechanistic interpretability}
 Mechanistic interpretability seeks to reverse-engineer neural networks into human-understandable algorithms, a critical challenge in trustworthy computer vision \cite{bereska2024mechanisticinterpretabilityaisafety}. Recent advances focus on decomposing model computations into interpretable subcomponents. Foundational approaches include Network Dissection \cite{bau2017networkdissectionquantifyinginterpretability}, quantitatively aligning convolutional filters with semantic concepts in datasets; as well as circuits \cite{Olah2020ZoomIA} as functional units in vision models, linking neurons to hierarchical visual features through activation visualization and ablation studies. For transformer-based architectures \cite{elhage2021mathematical} formalized attention head dynamics mathematically, revealing specialized roles in spatial and semantic processing. Another important work \cite{goh2021multimodal} identified "vision neurons" in CLIP that activate for abstract visual motifs, while \cite{cammarata2020curve} demonstrated how polysemantic neurons complicate interpretability in vision tasks. 
\section{Limitations}

Our study isolates mechanisms in a controlled setting, which necessarily introduces simplifying assumptions. We summarize key limitations and how they bound the scope of our conclusions.

\begin{enumerate}[label=(L\arabic*)]
\item \textbf{Uniform feature sparsity.}
We assume a fixed sparsity level across features (each feature is equally sparse). Real data rarely follow such homogeneity; activation frequencies are typically uneven or heavy–tailed. Extending the analysis to heterogeneous sparsity patterns would better reflect practical distributions.

\item \textbf{Disjoint task supports.}
Tasks are constructed from disjoint subsets of the feature space. This simplifies attribution but under–represents settings where tasks \emph{share} features. Our disjoint–support design is closest to scenarios where classes are cleanly separated along specific attributes (cf. fine–grained setups \cite{10.1145/3331184.3331336}), whereas in many applications a single feature can contribute to examples across multiple tasks. A controlled overlap parameter would allow interpolation between these extremes.

\item \textbf{Model class for dynamics (SINDy).}
We identify retention dynamics using a polynomial library of degree $\leq 10$. While expressive, this family cannot capture all possible relationships and risks misspecification for highly non–polynomial interactions. Library ablations or alternative function classes (e.g., rational terms, kernels, or neural bases) would test robustness of the learned relations.

\item \textbf{Retention metric vs.\ end performance.}
Our feature–level forgetting proxy tracks changes in representation strength, offering insight into where and when embeddings erode. However, it does not directly measure classifier performance. Establishing tighter links—e.g., systematic correlations with task accuracies and standard CL forgetting metrics—remains important future work.

\end{enumerate}

These limitations are by design: they trade realism for measurability. Future work will relax these assumptions (heterogeneous sparsity, shared/overlapping task features, richer dynamic families) and more directly connect retention signals to downstream accuracy and standard continual–learning metrics.

\section{Conclusions and future work}

\noindent \textbf{Conclusions.} 
This work examined how \emph{feature sparsity}, \emph{superposition}, and \emph{feature retention} interact in a controlled continual–learning (CL) setting. Our toy–world framework makes otherwise latent mechanisms observable: we probe single features through pure–feature inputs, quantify directional overlap, and track retention dynamics over training time, while also monitoring task–level capacity through effective rank. Three conclusions stand out. First, sparsity reliably drives overlap: when inputs activate fewer ground–truth features, the encoder economizes latent capacity via increased superposition, with a structured temporal profile—transient dips at task boundaries followed by consolidation–driven growth. Second, overlap is not uniformly harmful. The most vulnerable regime is \emph{low representation strength} under \emph{high overlap}; once a feature accumulates sufficient strength, its embedding remains comparatively stable even within a shared subspace. In short, forgetting depends on the interaction between overlap and strength rather than on overlap alone. Third, at the task level, sparser regimes yield higher effective ranks for task embeddings, indicating a broader allocation of latent directions to encode task structure despite greater feature–level sharing. Taken together, these findings refine the common intuition that “more superposition $\Rightarrow$ more forgetting,” and instead highlight representation strength and capacity allocation as primary moderators of retention in CL.

\noindent \textbf{Future works.} 
Several extensions follow naturally from our framework. A first step is to relax the disjoint–support design by introducing a tunable overlap parameter so that features can participate in multiple tasks; this more closely reflects realistic scenarios where classes share attributes and will test whether boundary dips and consolidation dynamics persist when interference is intrinsic to the task graph. We also plan to move beyond uniform sparsity by modeling heterogeneous, long–tailed feature frequencies, enabling analysis of how rarely used features accumulate strength, whether they suffer disproportionate forgetting, and how replay affects minority features over long horizons. To strengthen practical relevance, we will connect the retention proxy more directly to end performance by correlating feature–level dynamics with classwise accuracy and canonical CL forgetting measures, and by supplementing norm–based retention with alignment–based (directional) stability. On the modeling side, we will broaden the SINDy function library beyond degree–limited polynomials (e.g., rational terms, kernels, or neural bases). Finally, while our study intentionally uses a toy world for measurability, the same probes—pure–feature embeddings, overlap metrics, and effective–rank traces—can be layered onto richer architectures and datasets, creating a principled bridge from controlled mechanisms to realistic CL benchmarks.




{
    \small
    \bibliographystyle{plain}
    \bibliography{main}
}

\clearpage
\section{Appendix}

\subsection{SINDy}

Sparse Identification of Nonlinear Dynamical Systems (SINDy) is a data-driven method for discovering governing equations from time-series data. Given a system with state variables \( \mathbf{x}(t) \), SINDy assumes that their evolution follows a sparse set of nonlinear interactions:

\begin{equation}
\frac{d\mathbf{x}}{dt} = \mathbf{f}(\mathbf{x}).
\end{equation}

To identify \( \mathbf{f}(\mathbf{x}) \), SINDy constructs a library of candidate functions (e.g., polynomials, trigonometric terms) and applies sparse regression to select only the most relevant terms, enforcing model interpretability. The method is formulated as:

\begin{equation}
\dot{\mathbf{X}} = \Theta(\mathbf{X}) \Xi,
\end{equation}

where \( \dot{\mathbf{X}} \) represents the time derivatives of the system states, \( \Theta(\mathbf{X}) \) is a library of nonlinear candidate functions, and \( \Xi \) is a matrix of sparse coefficients indicating which terms are active. By iteratively refining \( \Xi \) using thresholded least squares, SINDy finds a compact mathematical representation of the system dynamics.

In this study, SINDy is applied to learn differential equations that govern the forgetting dynamics of features in continual learning, relating representation strength, superposition, and time since training.

\subsection{Results for different settings}
In this section, we present results for different settings. The details of the architectures can be found in tables \ref{tab:components2} \ref{tab:components3}. 
\begin{table}[h]
    \centering
    \renewcommand{\arraystretch}{1.2}
    \begin{tabular}{p{2cm} p{5.5cm}}
        \toprule
        Component & Architecture \\
        \midrule
        Generator & 
        \texttt{ReLU(Linear(50, 1500)),} \newline
        \texttt{ReLU(Linear(1500, 1500),} \newline
        \texttt{Linear(1500, 2000)} \\
        
        Separator &  \texttt{Softmax(Linear(50, 25))} \\  
        Encoder &  
        \texttt{ReLU(Linear(2000, 1500)),} \newline
        \texttt{ReLU(Linear(1500, 1500),} \newline
        \texttt{Linear(1500, 9)}\\  
        Classifier &  
        \texttt{Softmax(Linear(9, 25)),} \\
        \bottomrule
    \end{tabular}
    \caption{Architectures of the components of the controlled framework - setting 2}
    \label{tab:components2}
\end{table}

\begin{table}[h]
    \centering
    \renewcommand{\arraystretch}{1.2}
    \begin{tabular}{p{2cm} p{5.5cm}}
        \toprule
        Component & Architecture \\
        \midrule
        Generator & 
        \texttt{ReLU(Linear(50, 1500)),} \newline
        \texttt{ReLU(Linear(1500, 1500),} \newline
        \texttt{Linear(1500, 2000)} \\
        
        Separator &  \texttt{Softmax(Linear(50, 25))} \\  
        Encoder &  
        \texttt{ReLU(Linear(2000, 1500)),} \newline
        \texttt{ReLU(Linear(1500, 1500),} \newline
        \texttt{Linear(1500, 5)}\\  
        Classifier &  
        \texttt{Softmax(Linear(5, 25)),} \\
        \bottomrule
    \end{tabular}
    \caption{Architectures of the components of the controlled framework - setting 3}
    \label{tab:components3}
\end{table}

\begin{figure}[ht]
    \centering
\includegraphics[width=0.9\linewidth]{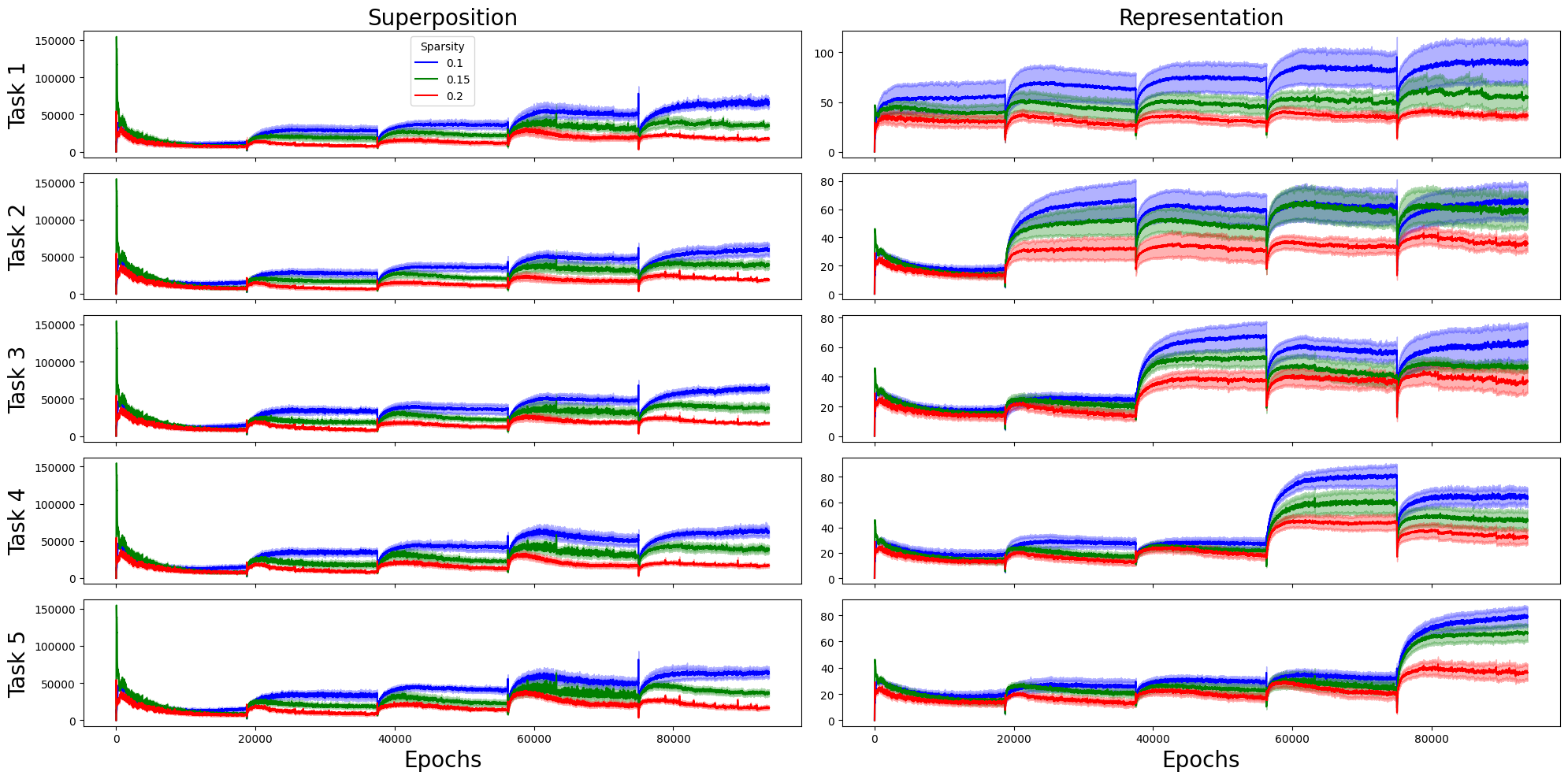}
    \caption{Dynamics of superposition and representation in time for setting 2}
    \label{fig:cumulative_superposition-setting2}
\end{figure}

\begin{figure}[ht]
    \centering
\includegraphics[width=0.9\linewidth]{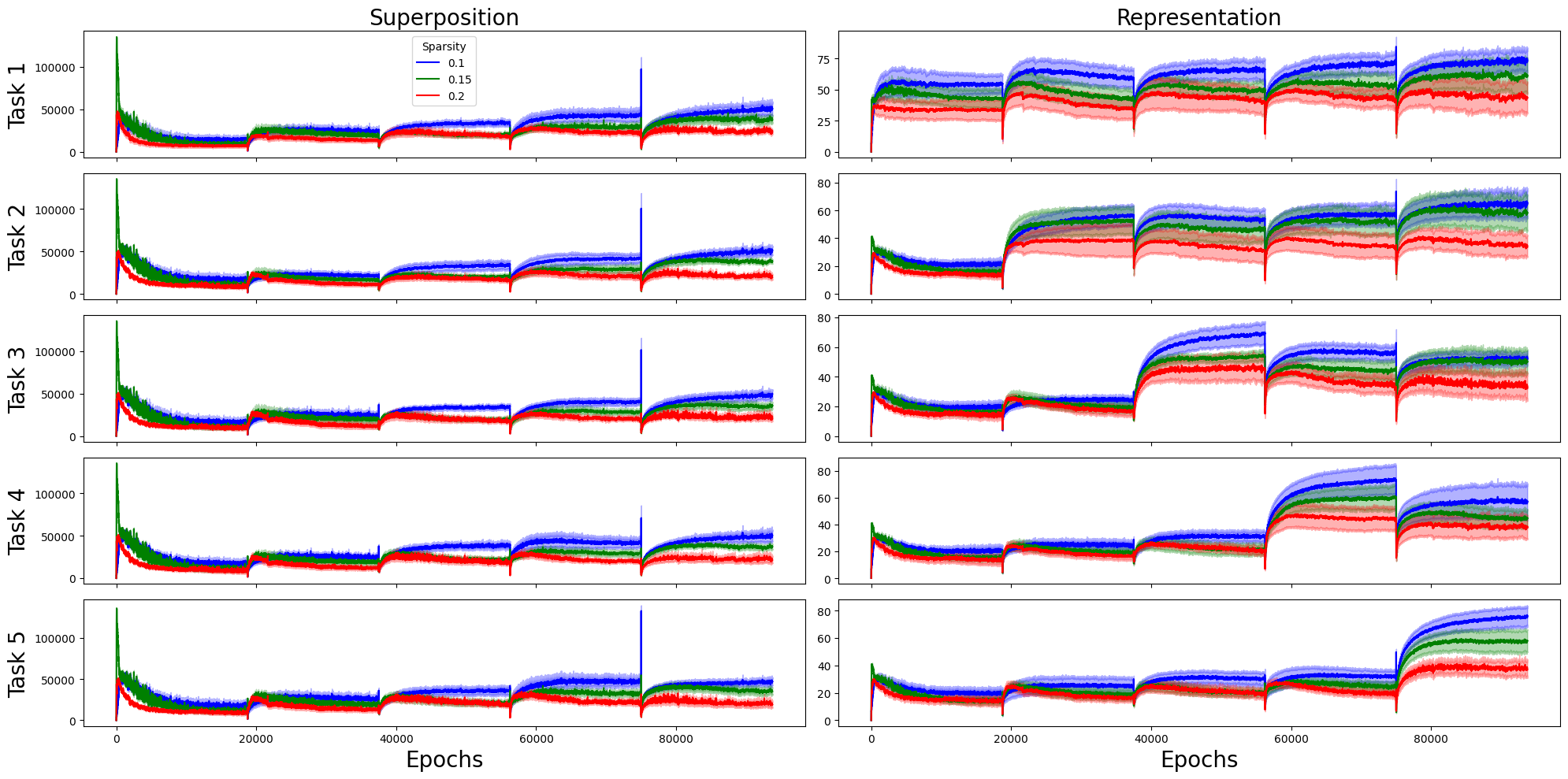}
    \caption{Dynamics of superposition and representation in time for setting 3}
    \label{fig:cumulative_superposition-setting3}
\end{figure}

\begin{figure}[ht]
    \centering
\includegraphics[width=0.9\linewidth]{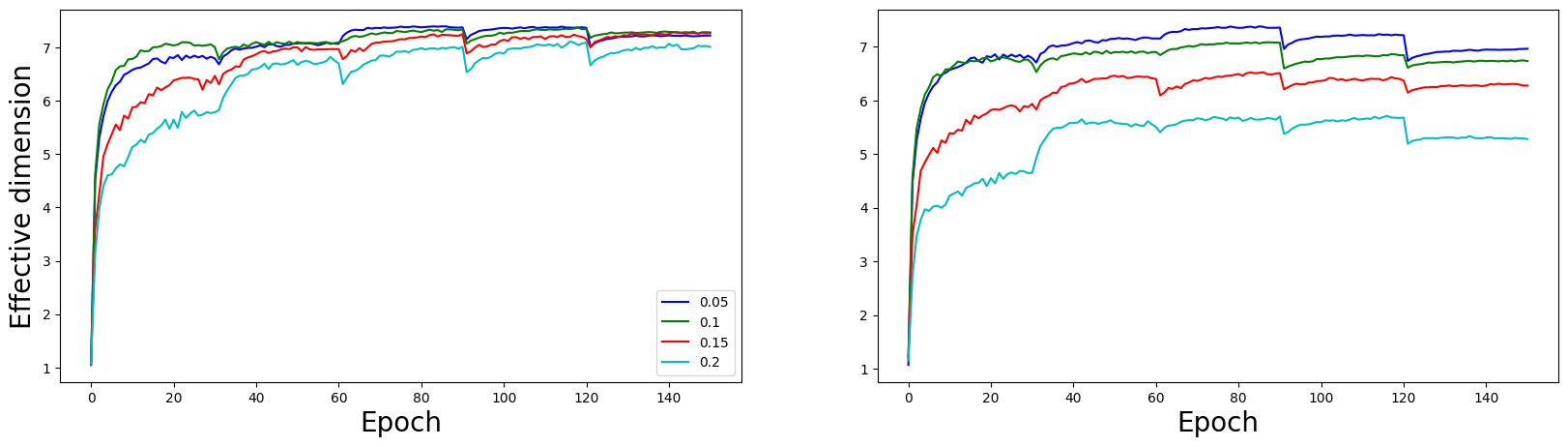}
    \caption{Effective dimensions of tasks vs sparsity-settings 2}
    \label{fig:sum_eff_dim2}
\end{figure}

\begin{figure}[ht]
    \centering
\includegraphics[width=0.9\linewidth]{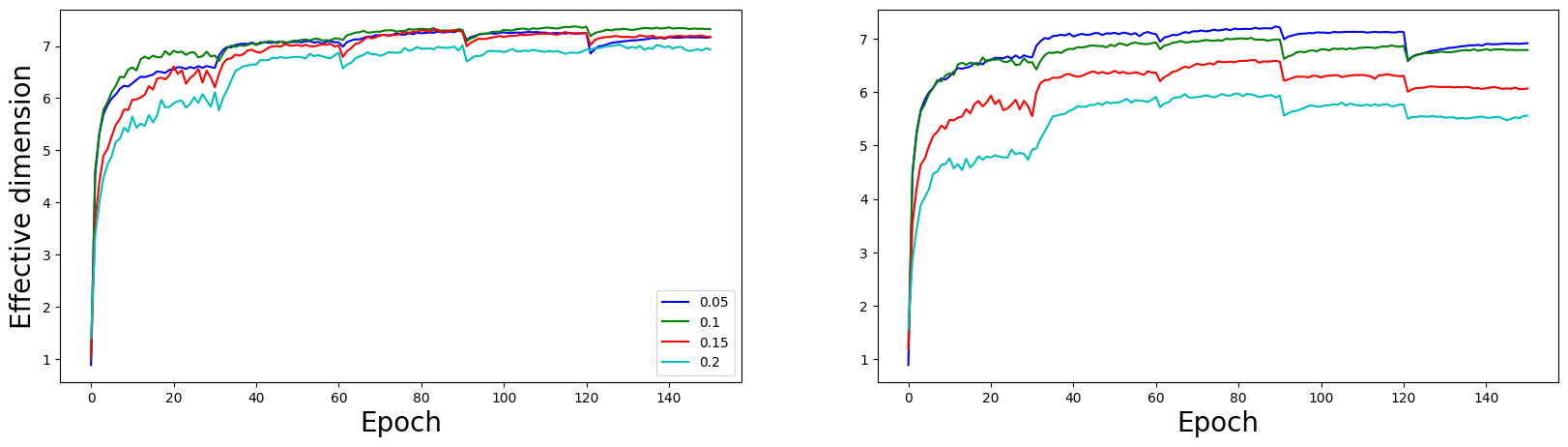}
    \caption{Effective dimensions of tasks vs sparsity-settings 3}
    \label{fig:sum_eff_dim3}
\end{figure}

\begin{figure*}[htb]
  \centering
  \includegraphics[width=\textwidth]{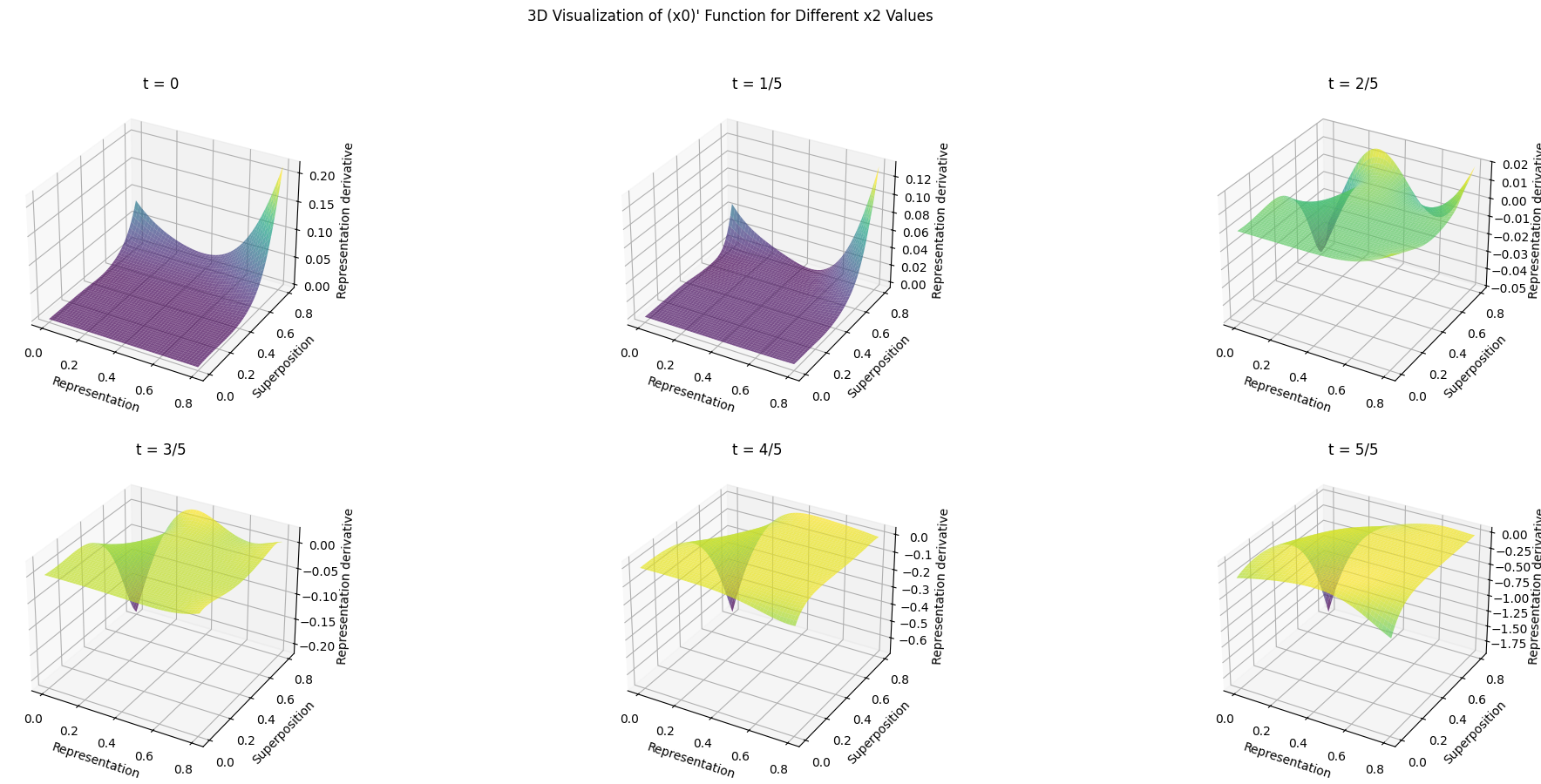}
  \caption{Instability of the method.}
  \label{fig:sindy}
\end{figure*}

\begin{figure*}[htb]
  \centering
  \includegraphics[width=\textwidth]{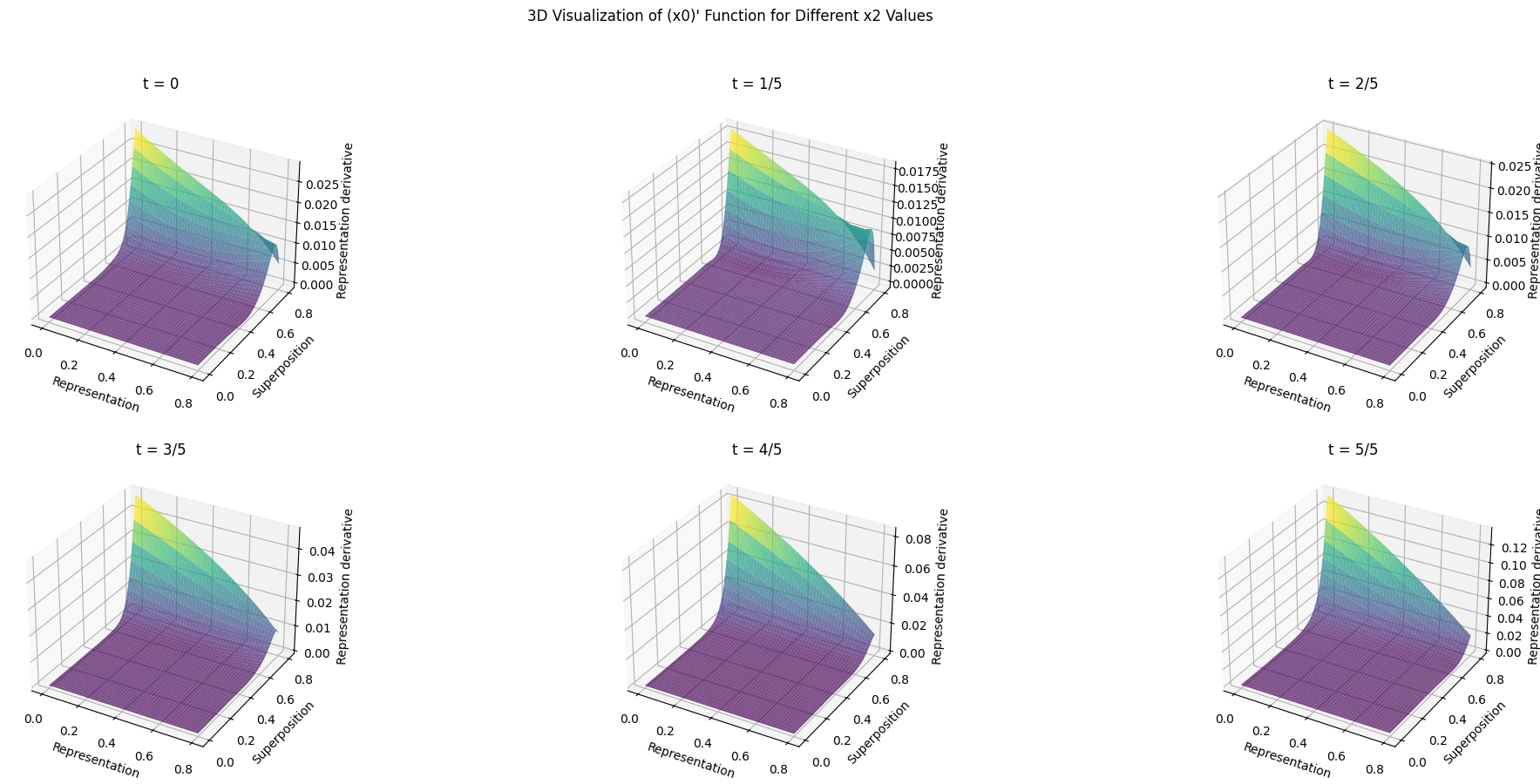}
  \caption{Instability of the method.}
  \label{fig:sindy}
\end{figure*}

\end{document}